# NLP in FinTech Applications: Past, Present and Future

Chung-Chi Chen,[1] Hen-Hsen Huang,[2,3] Hsin-Hsi Chen[1,3]

[1] Department of Computer Science and Information Engineering
National Taiwan University, Taiwan
[2] Department of Computer Science, National Chengchi University, Taiwan
[3] MOST Joint Research Center for AI Technology and All Vista Healthcare, Taiwan
cjchen@nlg.csie.ntu.edu.tw, hhhuang@nccu.edu.tw,
hhchen@ntu.edu.tw

## Abstract

Financial Technology (FinTech) is one of the worldwide rapidly-rising topics in the past five years according to the statistics of FinTech from Google Trends. In this position paper, we focus on the researches applying natural language processing (NLP) technologies in the finance domain. Our goal is to indicate the position we are now and provide the blueprint for future researches. We go through the application scenarios from three aspects including Know Your Customer (KYC), Know Your Product (KYP), and Satisfy Your Customer (SYC). Both formal documents and informal textual data are analyzed to understand corporate customers and personal customers. Furthermore, we talk over how to dynamically update the features of products from the prospect and the risk points of view. Finally, we discuss satisfying the customers in both B2C and C2C business models. After summarizing the past and the recent challenges, we highlight several promising future research directions in the trend of FinTech and the open finance tendency.

## 1 Introduction

Traditionally, financial service is highly regulated by the government because it influences everyone's daily life. In such a situation, only financial institutions such as commercial banks and investment banks can provide the services. In the recent trend of financial technology (FinTech), the situation is dramatically changed. Regulations are released, and the individual companies and the startups are allowed to provide financial services to the masses. Several disruptive innovations like P2P lending are emerging. In the financial revolution era, many technologies are used for overcoming the shortcomings of traditional financial services.

In the last five years, FinTech is one of the worldwide rapidly-rising topics. Figure 1 shows the statistics of "FinTech" from Google Trends. On the industrial side, there are many FinTech exhibitions held by enterprises and governments such as the Singapore FinTech Festival and the Fintech Week in Hong Kong and Canada[1]. On the academic side, many finance-related workshops are collocated with top-tier conferences, including ECONLP (Hahn et al., 2019, 2018) in EMNLP-2019 and ACL-2018, FinNLP (Chen et al., 2019c) in IJCAI-2019, and DSMM (Burdick et al., 2019) in CIKM-2019. Furthermore, FinTech is the theme of the 29th International Joint Conference on Artificial Intelligence, i.e., IJCAI-2020. All the events evidence the importance of FinTech development.

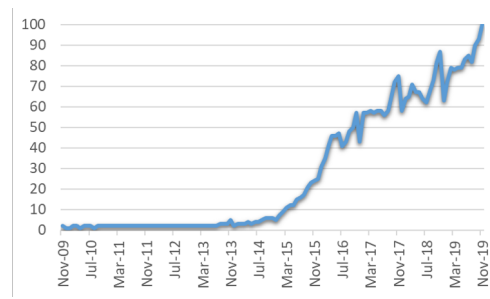

Figure 1: Statistics of "FinTech" in Google Trends.

With the recent progress of natural language processing (NLP), researchers start to pay more attention to tackle the unstructured data in the financial domain. In this position paper, we will discuss the past and the recent works that leverage NLP technologies to solve the problems in the financial domain or improve the financial services, and further blueprint the future research directions. Different from the previous overviews focusing on traditional financial research issues (Fisher et al., 2016) and the vanilla machine learning approaches (Das et al., 2014), we aim at the recent FinTech applica-tions and the development of state-of-the-art NLP methodologies. In particular, this paper is focused on the works from 2016 to 2019. Please refer to the previous survey paper for the works prior to 2016 (Fisher et al., 2016).

---
[1] https://www.fintechfestival.sg/; http://www.hongkong-fintech.hk/; http://www.fintechweek.ca/

We follow the main concepts in the financial industry, including know your customer, know your product, and satisfy your customer, to organize this paper. In each section, we introduce the issues by either the information sources or the E-commerce business models. For each application scenario, we review previous work, summarize the main idea of different methods, and provide the perspectives for future research directions.

The contributions of this paper are threefold as follows.

1. We provide an up-to-date survey focusing on NLP in the recent FinTech trend.

2. We integrate the researches and insights from both the NLP and the finance communities.

3. This paper presents a research agenda with several unexplored research directions for future works.

## 2 Know Your Customer (KYC)

As a highly-regulated industry, financial institutes are asked to evaluate their customers, including legal persons and natural persons, from different aspects such as identification and credit evaluation. In addition to using the structured data from the regular forms, a variety of textual data can be used to know the customers in depth. In this section, we separate the customers into corporate customers and personal customers.

### 2.1 Corporate Customers

Formal documents such as news articles and financial statements are mainly used to rapidly update the information of the corporate customers. For example, the news of financing pledge may influence the debt-paying ability of the company. In order to extract such financial events from financial announcements, Zheng et al. (2019) construct an end-to-end model with transformer encoder (Vaswani et al., 2017) and the BiLSTM-CRF event recognizer (Huang et al., 2015).

The operation situation can also be a cue for evaluating corporate customers. Zhai and Zhang (2019) use the information in 8-K reports with a sequence-to-sequence model to predict the material event of the firm.

Capturing the interactions between customers is also useful for understanding their operations. Oral et al. (2019) propose an algorithm that automatically constructs the relation graph from banking orders. Sakaji et al. (2019) use both news and bank contact histories to capture the relation between corporate customers with Granger causality analysis (Guo et al., 2008).

Legal issues may significantly hurt the development of the companies. Therefore, predicting the possible lawsuits the corporate customers may face is one of the important issues for financial institutions. Mao et al. (2019) propose a step-wise model with court announcement information to tackle this challenge, and their model performs the best in International Big Data Analysis Competition in IEEE ISI Conference 2019. Anti-money laundering (AML) is one of the important legal issues for financial institutions. Chen et al. (2019f) design a system that makes the checking process more efficient.

The information from online forums, blogs, and social media platforms are considered as the informal textual data. Such information can be used to capture the reputation of the brand or predict the sales of an enterprise. Lin et al. (2019) monitor the social media sentiment to predict the sales of the company with model-agnostic meta-learn method (Finn et al., 2017).

### 2.2 Personal Customers

With the flourishing social media platforms, using the personal daily posts to track the lifelogs of natural persons becomes possible (Yen et al., 2019). This kind of information is helpful for financial institutions, especially for insurance companies, to rapidly update the situation of their customers. The early detection or evaluation of the tendency of diseases (Losada et al., 2018, 2019) is an important issue for insurance companies when facing personal customers. For example, insurance companies can encourage and support their customers to get treatment early. The early detection of the diseases greatly increases the chances of successful treatment. Burdisso et al. (2019b) use SS3 method (Burdisso et al., 2019a) to overcome the early detection challenges.

Furthermore, with the record of social media posts, financial institutions can capture the social stratification of the new customers in a faster manner. Basile et al. (2019) use the stylistic information of the restaurant reviews with the convolutional neural network architecture (LeCun et al., 1995) to predict the social stratification of the writer.

## 2.3 Future Research Directions

Constructing the personal knowledge graph (Balog and Kenter, 2019) is one of the probable directions. The personal knowledge graph, which provides extra features from the customers' daily lifelogs, can be used in many scenarios, including the risk evaluation of insurance companies, the default possibility measurement of commercial banks, and the personalized precision marketing. It can also contribute to fraud detection (Wang et al., 2019).

In the open finance tendency, accessing the customers' transaction records in different financial institutions become possible. How to in-depth understand the customers and provide better service is an open challenge. As mentioned in Zibriczky (2016), there are few personalized stock recommendation systems (Chen et al., 2019a), and many existing stock recommendations systems do not consider the textual data (Tsai et al., 2019). In the future, constructing a recommendation system that can capture the personal behaviours is one of the major research directions.

## 3 Know Your Product (KYP)

Traditionally, KYP is a basic requirement for the salespersons in financial institutions. They must understand the attributes of the financial instruments they plan to merchandise to their customers. In this section, we broaden the concept of KYP to update the features of the products such as the prospect and the risk.

## 3.1 Prospect

Many works attempt to capture the price movement of the financial instruments. Some of these researches construct an end-to-end model for making the prediction. Hu et al. (2018) design a hybrid attention network (HAN) for predicting the stock trend with news. Yang et al. (2019) use bidirectional encoder representations from transformers (BERT) (Devlin et al., 2019) to encode the textual data related to the fear index (Engelberg and Gao, 2011), and experiment on S&P 500 index movement prediction. Chen et al. (2019d) leverage both BERT and the HAN model, and experiment on the foreign exchange market.

Some of researches extract useful information from the textual data. Keith and Stent (2019) extract the pragmatic and semantic features from earning calls to capture the analysts' decisions toward the target company. Chen et al. (2019e) use the extracted fine-grained events listed in the TOPIX finance event dictionary to make the stock price prediction. Ma et al. (2019) adopt Node2Vec (Grover and Leskovec, 2016) to construct news embeddings, and use the embeddings to predict the stock movement. Ding et al. (2019) predict the movement of the S&P 500 index by taking the intent and the sentiment information to account.

The information from the crowd has been shown useful for capturing the price movement since 2011 (Bollen et al., 2011). Chen et al. (2018) provide a fine-grained taxonomy for mining the opinions beyond sentiment from the financial social media users, and show that the information is comparable to professional analysts. To predict the stock price with financial tweets, Xu and Cohen (2018) construct a VAE-based (Semeniuta et al., 2017) end-to-end model, and Liu et al. (2019) propose a transformer-based (Vaswani et al., 2017) capsule network architecture (Sabour et al., 2017).

## 3.2 Risk

Risk is also an important attribute of financial instruments. Theil et al. (2018) propose a dictionary by word-embeddings to detect the uncertainty, and show a positive statistical relation between the uncertainty in the 10-K report and the stock volatility. Theil et al. (2019) combine the textual features in earning call and the financial features to predict the volatility of the stock. Qin and Yang (2019) use both verbal and vocal records from the conference calls to predict the risk of the companies with contextual BiLSTM architecture (Poria et al., 2017). Du et al. (2019) propose a system to evaluate the risk of the company from the financial reports.

## 3.3 Opportunities

Explainability is one of the open issues in the AI filed and even more important in the financial industry. Before selling financial instruments to customers, salespersons need to explain the rationales behind the products or the decisions. Izumi and Sakaji (2019) propose a demonstration to search the causal-chain from news. Learning to explain the market information as the reporters (Murakami et al., 2017) or the analysts is one of the possible research directions.

Numeral information is quite important when analyzing financial data. Extracting the numeral information and linking the relation between the numeral and the other named entities are very useful for financial textual data understanding (Lamm et al., 2018). Learning the sense toward numerals can be used to detect exaggerated information (Chen et al., 2019b). Bridging the numerals in textual data and the numerals in the table is also an important issue for financial textual data (Ibrahim et al., 2019). Because there are many numeral information in financial documents, tailor-made methods should be designed for dealing with the numerals.

## 4 Satisfy Your Customer (SYC)

Many startups are springing up to share the market of traditional financial institutions. In the FinTech industry, developers and researchers attempt to make the financial activities more efficient and more liberal. "Satisfy your customers" (SYC) becomes a new focus of the financial institutions. In the FinTech revolution, people pay close attention to leverage technology to satisfy those customers, not in the VIP-class. In this section, we classify the works by their business models, say, Business to Customer (B2C) and Customer to Customer (C2C).

### 4.1 Business to Customer (B2C)

Constructing a dialog system that supports customer service is a recent tendency in the service industry. As a special service industry, financial institutions are no exception. There are some fundamental researches related to the development of the financial dialog system. Lai et al. (2018) propose a BiLSTM-based model for product-related question answering. Altinok (2018) propose an ontology-based dialogue management system, and Rivera et al. (2019) provide a dataset with dialog act labels for question answering.

Recommendation systems have been adopted in many different domains in finance such as banking, insurance, and so on. Zibriczky (2016) provide a literature review for these systems. Here, we focus on the recent recommendation systems using textual data. Sun et al. (2018) adopt the sentiment analysis results on the social media platform for the stock recommendation. By taking the posts on the financial social media platform into consideration, Chen et al. (2019a) predict the personalized next-interested stocks with a joint learning model.

### 4.2 Customer to Customer (C2C)

More and more transactions and information exchange are directly done in the person-to-person business model. For the platform provider, this business model can be seen as the customer-to-customer model. Peer-to-peer (P2P) lending is one of the famous functions, while risk evaluation is an extremely important issue. Li et al. (2019) predict the intermediary risk with profile textual data via feature extraction models.

### 4.3 Challenges

The lack of publicly available datasets is one of the big issues for the researchers who focus on both NLP and finance. For instance, how to construct a multi-term dialogue system is an open challenge in the NLP fields. One of the important issues is that automatically classifying the intents of the customers from the first few terms. To the best of our knowledge, however, no dataset is publicly available for this task in the financial domain.

Nowadays, amateurs can easily share the market of traditional firms. For example, YouTubers have influenced the entertainment industry. The social trading platforms also provide the place for individual investors to share the market of professional analysts. How to evaluate the performances and the opinions of the users in social trading platforms remains an important issue (Ying and Duboue, 2019).

The order or chart from doctors can not only be used for the ICD code prediction task (Xie and Xing, 2018; Bai and Vucetic, 2019), but can also be used for the insurance industry. The insurance companies settle insurance claims based on these textual data. Therefore, automatically understanding the clinical documents is also useful for making the settle process more efficient.

## 5 Conclusion

FinTech is an emerging area in which many attempts have been explored since 2015. In this paper, we provide an overview of the applications and the related approaches in the FinTech trend, and blueprint promising future research directions for NLP and finance researchers. We hope that this position paper can inspire interdisciplinary researchers to focus on this topic and set the cornerstone for future research.